\documentclass{article} 
\usepackage{iclr2026_conference,times}

\usepackage{amsmath,amsfonts,bm}

\def\eqref#1{equation~\ref{#1}}

\def\1{\bm{1}}

\DeclareMathAlphabet{\mathsfit}{\encodingdefault}{\sfdefault}{m}{sl}
\SetMathAlphabet{\mathsfit}{bold}{\encodingdefault}{\sfdefault}{bx}{n}

\usepackage{hyperref}
\usepackage{url}
\usepackage{graphicx}
\usepackage{multirow}
\usepackage{multicol}
\usepackage{booktabs, multirow} 
\usepackage{soul}
\usepackage{float}
\usepackage{xcolor,colortbl} 
\usepackage{changepage,threeparttable} 
\usepackage{subcaption}
\title{An Exploratory Study of Blood Glucose Estimation from Photoplethysmography Signals using Machine Learning}

\author{Ruhani Bhatia, Vijval Ekbote \\
Indraprastha Institute of Information Technology, Delhi\\
}

\iclrfinalcopy 
\begin{document}

\maketitle

\begin{abstract}
Diabetes and extreme blood sugar levels are some of the major health problems faced by humans today across the world. While Continuous Glucose Monitoring (CGM) has emerged as an effective technology for management of diabetes as well as for monitoring blood sugar levels, this technology has traditionally been invasive (that is, requiring the piercing of the skin) and carries the risk of irritation, induration, etc. This highlights the need for accurate and non-invasive CGM methods that can be deployed at scale. With the emergence of various sensing technologies and their integration in wearables like the smart-watch, we now have the capability to continuously monitor body signals like the Photoplethysmogram (PPG) in a non-invasive manner. Having the ability to continuously monitor blood glucose through CGMs and continuously monitor PPG signals through a smart-watch offers an opportunity to get dense data on these two, opening the possibility of building machine learning and deep learning based models to estimate blood glucose level from PPG signals. In this work, we first present a paired dataset comprising continuous PPG signals from a smartwatch along with glucose values recorded using a CGM device. We also present the results of some preliminary experimental explorations performed on our dataset. These preliminary results suggest that some predictive signals may exist, though more exploration is needed with more data from a larger number of individuals. The dataset can be accessed at \href{https://zenodo.org/records/20577959}{this link}
\end{abstract}

\section{\bf Introduction}

Diabetes is one of the leading causes of death worldwide, and its presence leads to a higher risk of mortality from cardiovascular diseases, stroke, and chronic liver disease, among other ailments \citep{lin2020global}. It also has the second largest negative effect on global health adjusted life expectancy, and the risk of all-cause mortality is 2-3 times more for diabetes patients \citep{lin2020global}. Given the risks posed by diabetes, its management becomes all the more important. One of the most popular methods for the same is continuous glucose monitoring (CGM), which involves continuously measuring blood glucose levels, typically using a device which includes a sensor placed under the skin (and is hence invasive), and thus may pose a risk of infections and skin-irritation. There is thus a need for other alternatives which are durable and non-invasive, while retaining accuracy and performance.

The rise of machine learning (ML) and deep learning (DL) have played an important role in the creation of many new applications, mainly due to their ability to approximate complicated functions, and have thus found their way into many new domains. One such domain of interest is the prediction of quantities of clinical interest by using ML systems which accept common physiological signals such as Electroencephalogram (EEG) and Electrocardiogram (ECG) as inputs.  In particular, various deep learning methods have been used in the past to process EEG signals for tasks such as emotion recognition and disease detection \citep{li2016emotion, kim2023deep}, and work has also been done on applying DL methods to ECG signals for detection of cardiovascular diseases \citep{iqbal2024novel}. 

For our study, we use Photoplethysmography (PPG), which is a non-invasive method to measure variations in peripheral blood volume by using low-intensity Infrared light. The intensity of this light varies due to its absorption by blood, and these changes in intensity are used to measure blood volume changes. PPG is often used in clinical settings to measure blood oxygen saturation (SpO2). Recognising the potential utility of these signals as viable and practical inputs for clinical ML systems, previous works have experimented with a number of different applications and methods. \citep{srinivasan2021deep}, for instance, make use of the MIMIC-III dataset for the task of diabetes prediction, training convolutional neural networks (CNNs) on scalograms obtained from PPG signals. \citep{islam2021blood} make use of videos captured using smartphone cameras to derive PPG signals, which were subsequently passed to classical ML models for prediction of blood glucose level. In \citep{satter2024emd}, features derived from the signals using methods like Empirical Mode Decomposition were finally passed to ML models for estimation of glucose values. However, in all these works, glucose levels were not recorded for long durations of time (most works recorded values for a maximum of 3 minutes).

To exploit the practicality and non-invasive nature of PPG signals, we propose to develop a machine learning-based continous blood glucose monitoring system using PPG signals as input. A key challenge in building such systems is the lack of dense data that pairs PPG signals with the corresponding blood glucose values. Previous works have collected such data for short durations and not over extended periods of time \citep{satter2024emd, islam2021blood}. In order to overcome these challenges, we first collect dense continuous data in the form of PPG signals paired with the corresponding blood glucose measurements using a CGM device to create a paired dataset. This data was collected over a two-week period. Numerous processing techniques and feature engineering methods were then applied on this data to prepare it for modeling, after which DL algorithms were utilised for final prediction of glucose levels. We explored numerous challenges such as interpolation of recorded glucose values, effective feature extraction and engineering, and assessed the generalizability of our models.

\section{\bf Method}
\subsection{\bf Dataset Creation}
For our experiments, we made use of a paired dataset created by collecting PPG data and blood glucose levels simultaneously from a total of 5 volunteers, with a male:female ratio of 3:2. Each volunteer wore an \href{https://www.empatica.com/embraceplus}{Empatica EmbracePlus} Smartwatch for 14 days. The wristwatch provided us with raw PPG signal values, with a sampling frequency of 64 Hz. Additionally, each volunteer also wore a CGM device (\href{https://www.freestyle.abbott/us-en/home.html}{Freestyle Libre}) for the same duration of 14 days. We thus obtained PPG and glucose data (in mmol/L) simultaneously for all the volunteers. The data collected from the smartwatch was stored in the cloud using Amazon Web Services (AWS) in the avro file format, and the data from the CGM device was obtained as a csv file at the end of the stipulated period. Before collecting data for this study, due permission was obtained from the Institute review board. The dataset can be accessed at \href{https://zenodo.org/records/20577959}{this link}.

\subsection{\bf Feature Extraction and Engineering}
\subsubsection{Interpolation}
The CGM device we used collected glucose data at 15 minute intervals, while the smartwatch provided PPG signal values at a sampling rate of 64 Hz (64 values per second). This disparity in the density of the two types of data would lead to underutilisation of the PPG data, since the number of data points we could use for training would be restricted by the number of glucose values. To overcome this disparity, we used cubic spline interpolation to fit a smooth curve to the glucose values. Obtaining this smooth curve enabled us to sample glucose values at any timestep we required leading to a much larger number of datapoints for training.

\subsubsection{Feature Extraction}
Training models on raw PPG values is impractical due to the high sampling rate and possibility of noisy measurements. We thus extracted features via the pyPPG library using a sliding window framework, since this would give us a fixed number of features regardless of window size, and be less influenced by noise as compared to raw values. A total of 918 features (encompassing temporal, spectral, and morphological) were derived from the raw signal sampled at 64 Hz. For this analysis, a 60-second window duration was implemented to ensure the inclusion of multiple complete cardiac cycles and to provide sufficient signal context.

Furthermore, a 50\% temporal overlap (30-second stride) was applied between successive windows. This sliding window approach serves two critical functions: first, it provides a smoother transition in the feature space, preventing the loss of transient physiological events that might be bisected by rigid window boundaries; second, it effectively doubles the available data density for modeling, without the computational overhead of a denser stride.

We finally obtain a 918-dimension feature vector for each sixty second window of PPG data. Each of these feature vectors is paired with a glucose value sampled from the interpolated cubic spline obtained previously. The glucose value is taken from the timestep at the end of each sliding window.

Some of the features extracted were as follows:
\begin{itemize}
    \item \textbf{Time domain features}: Features such as maximum amplitude of the signal, mean amplitude, skewness, kurtosis, energy over a time window, etc.
    \item \textbf{Frequency domain features}: dominant frequency in the signal, total power, etc
    \item \textbf{Features related to Heart Rate Variability}: This category includes features such as mean inter-beat interval, standard deviation of normal-to-normal intervals, beats per minute, etc.
\end{itemize}

\section{\bf Experiments and Results}

We conducted a number of preliminary experiments on smaller subsets of our data in order to probe interesting directions for future research. We present our findings in this section.

\subsection{Training and Testing on the same subject}
This is our base configuration. We perform a temporal 80-20 train-test split (the initial 80\% of the data is used for training, and the subsequent 20\% for testing), and use a multi layered perceptron (MLP) as our model. The MLP had two hidden layers, each with 100 neurons, and used the Rectified Linear Unit (ReLU) as the activation function to add non-linearity. The network was optimized using the Adam optimizer \citep{kingma2014adam} with a learning rate of 0.001.  To assess performance, we employ the Mean Squared Error (MSE) and the Mean Absolute Percentage Error (MAPE). The MSE measures the average squared difference between the predicted glucose values and the corresponding ground truth values, while the MAPE measures the average absolute prediction error expressed as a percentage of the true glucose values. We report the day-wise results (for four different days) for one of the subjects, in table \ref{tab:mse_mape_same_person}

\begin{table}[t]
\centering
\caption{Predictive performance on the same subject, on four different days}
\label{tab:mse_mape_same_person}
\begin{tabular}{c c c}
\hline
\textbf{Day Number} & \textbf{MSE} & \textbf{MAPE (\%)} \\
\hline
Day 1 & 1.04 & 14.28 \\
Day 2 & 0.64 & 12.08 \\
Day 3 & 0.48 & 10.26 \\
Day 4 & 0.46 & 7.65 \\
Avg & 0.65 & 11.06
\end{tabular}
\end{table}

\subsection{Personalizing the trained model for a particular subject}
We conduct an experiment to evaluate if a general trained model can be personalized for a particular subject. This is especially useful for CGM due to variability in glucose values among people belonging to different demographics.
We trained the MLP on three subjects' data, then finetuned it on the data of the fourth subject. The predictions were smoothed by maintaining the exponentially weighted moving average of the previous predictions. 
We measured the Root Mean Squared Error (RMSE), the Mean Absolute Error (MAE), and the Mean Error (ME). The model obtains an MAE of 1.0581, and an RMSE of 3.5597. 
The Mean Error (Bias) of -0.6850 reveals a slight systematic underestimation (negative bias), where the predicted values typically reside marginally below the ground truth. To assess how accurately our predictions are able to track the ground truth glucose values, we calculated Lin's Concordance Correlation Coefficient (CCC), which measures the agreement between two sets of continuous variables. CCC lies in the interval [-1,1], with 1 indicating perfect agreement, 0 indicating no agreement, and -1 denoting perfect disagreement. We observe a very low value of 0.0142, suggesting that despite relatively low RMSE, our predictions are not effectively synchronized with the ground truth glucose values. 

\subsection{Predicting Glucose Values with time lag}
We had conducted an experiment where we tried to predict the glucose value some $\tau$ minutes later in the future, rather than predicting the value for the current timestamp $t$. This was based on the assumption that the features obtained from the current window of the PPG signal may not be indicative of the immediate value of glucose; rather, it may take time for the changes reflected by these features to be observed in the glucose values. We conduct experiments using different values of $\tau$ (10, 15, and 20 minutes), and report the MSE and MAPE. In table \ref{tab:lag_results} we report the results (averaged over all days) for 2 subjects. We note that there is a significant difference in the performance on the 2 subjects, indicating that this approach may not be very stable.
\begin{table}[t]
\centering
\caption{Average predictive performance across different lag values ($\tau$).}
\label{tab:lag_results}
\begin{tabular}{c|cc|cc}
\hline
\multirow{2}{*}{$\tau$ (min)} & \multicolumn{2}{c|}{Subject 1} & \multicolumn{2}{c}{Subject 2} \\
 & MSE & MAPE (\%) & MSE & MAPE (\%) \\
\hline
10  & 0.80 & 12.12 & 3.3   & 15.63   \\
15 & 0.75 & 12.07 & 2.46 & 13.94 \\
20 & 0.88 & 12.74 & 2.83 & 15.45 \\
\hline
\end{tabular}
\end{table}

\subsection{Clinical Safety Assessment}
To assess the performance of our fine-tuned model, we use the Clarke Error Grid to analyse the predictions. Often considered a good standard for the evaluation of the accuracy of blood glucose measuring devices, this method divides a scatter plot of the predicted values into different regions based on the deviations from reference values. Predicted points lying in Region A differ from the reference values in magnitude by no more than 20\%. Predictions lying in this region are considered to be the most accurate. Another standard that is used to measure reliability of CGMs is the ISO (International Organisation for Standardisation) 15197:2013. According to this standard, for a glucose measuring device to be approved for production, atleast 95\% of its predictions must have a magnitude within ±15\% of the corresponding reference glucose levels in case the reference is \(\geq 100 \, \text{mg/dl}\), and within ±15 mg/dl in case the reference value is \(< 100 \, \text{mg/dl}\) \citep{jendrike2017iso}.
We observe that our results are not yet satisfactory for clinical use, since only 73.64\% of our predicted data points fall within Zone A (clinically acceptable) and 25.63\% points fall in Zone E, which represents errors that may lead to wrong treatment. Therefore, we do not recommend clinical deployment without further validation. We report the full distribution in table \ref{tab:clarke_error_grid}

\begin{table}[h]
\centering
\caption{Clarke Error Grid Analysis Results}
\label{tab:clarke_error_grid}
\begin{tabular}{|c|c|p{8cm}|}
\hline
\textbf{Zone} & \textbf{Percentage (\%)} & \textbf{Clinical Meaning} \\
\hline
A & 73.64 & Clinically accurate values \\
\hline
B & 0.00 & Benign errors; no effect on clinical action \\
\hline
C & 0.00 & Errors leading to unnecessary treatment \\
\hline
D & 0.73 & Errors leading to failure to detect a dangerous state \\
\hline
E & 25.63 & Dangerous errors leading to incorrect treatment \\
\hline
\end{tabular}
\end{table}

\section{\bf Conclusion}
In this work, we investigated the challenge of non-invasive blood glucose level prediction by leveraging PPG signals obtained from a smart watch. We developed a dense paired dataset of PPG signals and corresponding glucose values collected through a Continuous Glucose Monitoring (CGM) device, which we hope will serve as a useful resource for future research in this area. Our experiments with a variety of different configurations suggest the potential of using wearable devices for non-invasive glucose monitoring, but extensive validation and clinical testing would be required before any claims about accessibility or accuracy can be made. We hope that this dataset, along with our findings, will inspire further research, and contribute to the scientific community in a positive way. 

\section{Acknowledgements}
We would like to sincerely thank Professor Pankaj Jalote who conceptualized and initiated this project,  got the data collected and guided us throughout the project. We would also like to thank students who worked on this project at different points in time, and contributed to exploration of data, methods, etc., namely - Harshit Gautam, Shivam Dwivedi, Sahil Sahu, and Subhanshu Bansal. 

\bibliography{iclr2026_conference}
\bibliographystyle{iclr2026_conference}

\appendix
\section{Appendix}
\subsection{Analysis of Interpolation Methods}
We present some observations we obtained using various interpolation methods, and try to empirically motivate our choice of using cubic spline interpolation. We choose 20 consecutive glucose values from one of the subjects, and show the plots obtained using Lagrange interpolation (with degree 19), polynomial interpolation (degree 19) and cubic spline interpolation. As can be observed, cubic spline interpolation gives the smoothest curve, with minimal fluctuations. In constrast, Lagrange interpolation shows severe fluctuations at the boundaries.

\begin{figure}[!htbp]
    \centering
    \includegraphics[width=0.85\textwidth]{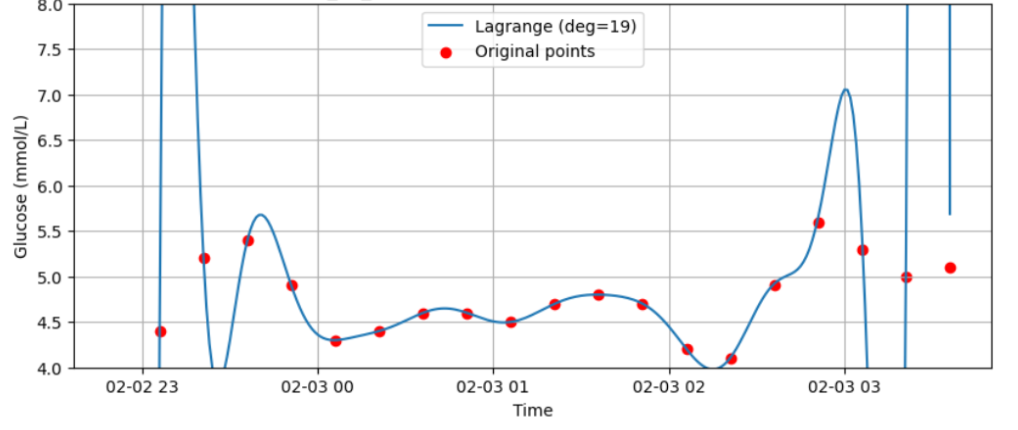}
    \caption{Curve obtained using Lagrange Interpolation}
    \label{fig:lagrange}
    \vspace{4pt}
    \includegraphics[width=0.85\textwidth]{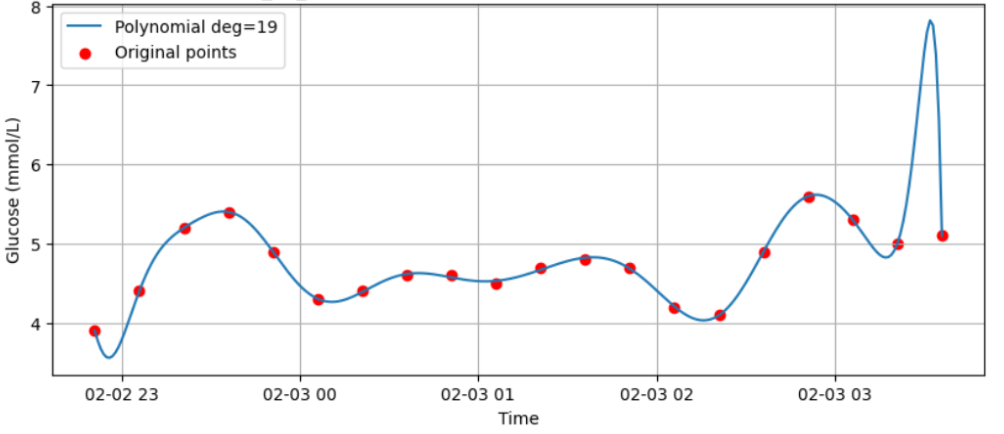}
    \caption{Curve obtained using Polynomial interpolation}
    \label{fig:poly}
    \vspace{4pt}
    \includegraphics[width=0.85\textwidth]{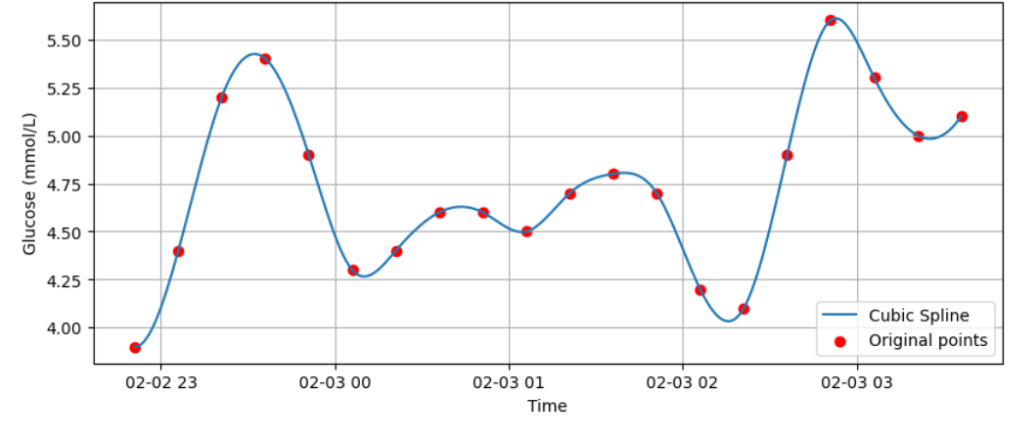}
    \caption{Curve obtained using Cubic-Spline interpolation}
    \label{fig:cubic}
\end{figure}

\clearpage
\subsection{Day-wise results for glucose prediction with time lag}
\begin{table}[H]\centering\small
\caption{Detailed results for glucose prediction with lag. We report results for 2 subjects, with three different values of $\tau$: 10, 15, and 20. Blank values indicate no data was available for that day.}\label{tab:lag_full}
\renewcommand{\arraystretch}{0.85}
\begin{tabular}{l|r|r|r|r|r|r|}\toprule
& &\multicolumn{2}{c|}{Subject 1} &\multicolumn{2}{c|}{Subject 2} \\\cmidrule{3-6}
$\tau$ &Day &MSE &MAPE(\%) &MSE &MAPE(\%) \\\midrule
\multirow{14}{*}{10} &Day 1 &2.4 &19.4 &0.85 &12.31 \\
&Day 2 &1.12 &15.91 &2.09 &13.9 \\
&Day 3 &0.72 &12.88 &12.33 &20.24 \\
&Day 4 &0.82 &15.14 &1.08 &13.2 \\
&Day 5 &1.01 &16.17 &9.33 &19.1 \\
&Day 6 &0.4 &11.03 &0.95 &14.84 \\
&Day 7 &1.65 &16 &1.46 &20.88 \\
&Day 8 &0.53 &8.87 &1.22 &18.84 \\
&Day 9 &0.3 &7.6 &1.41 &17.74 \\
&Day 10 &0.29 &8.1 &9.94 &15.87 \\
&Day 11 &0.28 &8.16 &1.15 &17.02 \\
&Day 12 &0.16 &6.27 &0.67 &12.52 \\
&Day 13 & & &0.42 &6.77 \\
&\textbf{Avg} &\textbf{0.80} &\textbf{12.12} &\textbf{3.3} &\textbf{15.63}
\\\cmidrule{1-6}
\multirow{14}{*}{15} &Day 1 &2.14 &17.86 &0.81 &11.05 \\
&Day 2 &1.06 &16.38 &1.23 &11.68 \\
&Day 3 &0.65 &12.18 &10.58 &17.65 \\
&Day 4 &0.72 &14.41 &1.18 &13.7 \\
&Day 5 &1 &16.68 &5.38 &15.69 \\
&Day 6 &0.53 &12.96 &0.83 &12.88 \\
&Day 7 &1.41 &15.13 &1.39 &19.87 \\
&Day 8 &0.51 &8.98 &0.93 &15.52 \\
&Day 9 &0.38 &8.03 &1.2 &15.49 \\
&Day 10 &0.3 &8.5 &6.64 &14.49 \\
&Day 11 &0.25 &8 &0.75 &13.42 \\
&Day 12 &0.14 &5.8 &0.52 &10.29 \\
&Day 13 & & &0.57 &9.57 \\
&\textbf{Avg} &\textbf{0.75} &\textbf{12.07} &\textbf{2.46} &\textbf{13.94} 
\\\cmidrule{1-6}
\multirow{14}{*}{20} &Day 1 &2.12 &17.46 &0.96 &12.44 \\
&Day 2 &1.12 &16.97 &0.93 &12 \\
&Day 3 &0.64 &11.95 &9.08 &19.33 \\
&Day 4 &0.81 &15.13 &1.26 &14.39 \\
&Day 5 &1.06 &17.35 &8.22 &19.7 \\
&Day 6 &0.6 &14.02 &1 &14.49 \\
&Day 7 &1.54 &16.3 &1.67 &22.7 \\
&Day 8 &0.51 &9.68 &1.28 &18.61 \\
&Day 9 &1.4 &10.3 &1.4 &17.68 \\
&Day 10 &0.35 &9.55 &9.13 &15.42 \\
&Day 11 &0.32 &8.62 &1.02 &16.14 \\
&Day 12 &0.14 &5.56 &0.64 &11.87 \\
&Day 13 & & &0.29 &6.1 \\
&\textbf{Avg} &\textbf{0.88} &\textbf{12.74} &\textbf{2.83} &\textbf{15.45} \\
\bottomrule
\end{tabular}
\end{table}

\end{document}